\documentclass[10pt,twocolumn,letterpaper]{article}

\usepackage{cvpr}
\usepackage{times}
\usepackage{epsfig}
\usepackage{graphicx}
\usepackage{amsmath}
\usepackage{amssymb}

\usepackage{mathrsfs} 
\usepackage{times,amstext,amsmath,amssymb,latexsym,shapepar,array,epsfig,bm,paralist}
\usepackage{amsfonts,subfigure}
\usepackage{amssymb}
\usepackage{amsthm,amsmath}
\usepackage{graphicx}
\graphicspath{{figs/}}
\newcommand{\beq}{\begin{equation}}
\newcommand{\eeq}{\end{equation}}
\newcommand{\beqs}{\begin{eqnarray}}
\newcommand{\eeqs}{\end{eqnarray}}
\newcommand{\barr}{\begin{array}}
\newcommand{\earr}{\end{array}}

\usepackage[breaklinks=true,bookmarks=false]{hyperref}

\cvprfinalcopy 

\def\cvprPaperID{553} 

\begin{document}

\title{Low-Cost Compressive Sensing for Color Video and Depth}

\author{Xin Yuan, Patrick Llull, Xuejun Liao, Jianbo Yang, Guillermo Sapiro, David J. Brady and Lawrence Carin\\
Department of Electrical and Computer Engineering, Duke University,
Durham, NC, 27708, USA\\
{\tt\small \{xin.yuan, prl12, xjliao, jy118, dbrady, guillermo.sapiro, lcarin\}@duke.edu}
}

\maketitle

\begin{abstract}
A simple and inexpensive (low-power and low-bandwidth) modification is made to a conventional off-the-shelf color video camera, from which we recover {multiple} color frames for each of the original measured frames, and each of the recovered frames can be focused at a different depth.
The recovery of multiple frames for each measured frame is made possible via high-speed coding, manifested via translation of a {single} coded aperture; the inexpensive translation is constituted by mounting the binary code on a piezoelectric device. To simultaneously recover depth information, a {liquid} lens is modulated at high speed, via a variable voltage. Consequently, during the aforementioned coding process, the liquid lens allows the camera to sweep the focus through multiple depths. 
In addition to designing and implementing the camera, fast recovery is achieved by  
an anytime algorithm {exploiting} the group-sparsity of wavelet/DCT coefficients.
\end{abstract}

 \begin{figure*} 
\begin{center}
   \includegraphics[scale=0.35]{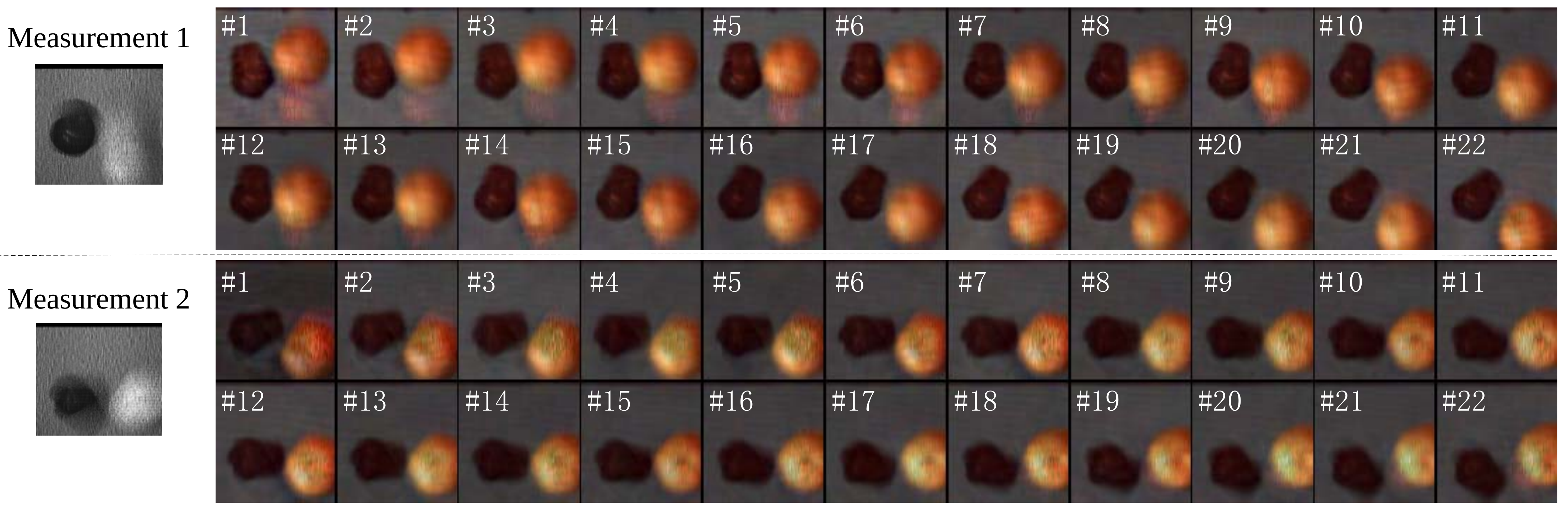}
\end{center}
\vspace{-0.3cm}
\caption{\small Reconstruction of real measurement with frame size $512\times 512$. Two plastic fruits dropping, touch a surface at the bottom, and then rebound. Left-most are two contiguous measurements, and the right part shows 44 corresponding reconstructed video frames ($n_t=22$ recovered video frames from each compressive measurement). We can see clear motion and color information from the reconstructed frames (video online at \cite{Website}).  Note in this dataset we did not change the focal plane during the capture of the measurement.}
\label{Fig:real_ball}
\vspace{-0.4cm}
\end{figure*}


\vspace{-5mm}
\section{Introduction}
\vspace{-3mm}

A variety of techniques have been developed to extract information from a single image.
For example, the depth-from-focus method \cite{Levin10,Zhou09a,Zhou11} allows estimation of a 3D scene (depth-dependent focus) from a single 2D image.
The mosaic and demosaic technique~\cite{Adams98} allows recovery of color information from a gray-scale image. 
Recently, inspired by compressive sensing (CS) \cite{cs_Candes06}, video has been extracted from a single image ~\cite{Hitomi11ICCV,HollowayICCP12,Patrick13OE,Reddy11CVPR}. In this setting the measured data are acquired at a low frame rate, with coding at a faster rate, and high-frame-rate video is {computationally recovered subsequently}.

In this paper we develop a new method that borrows and extends ideas from this previous work. Specifically, like \cite{Hitomi11ICCV,HollowayICCP12,Patrick13OE,Reddy11CVPR} we perform high-frequency coding of video collected at a low frame rate, with CS-based inversion. Our coding strategy differs from previous work in that we use a \emph{single} code that is inexpensively translated via a piezoelectric device. We recover color via a {hardware mosaicing and computational demosaicing} procedure like in conventional cameras. The newest aspect of the proposed approach is that we use a lens with voltage-dependent index of refraction (a liquid lens), and by varying the voltage at high rate, the recovered high-rate video corresponds to capturing data at varying focus points (depths). For each of the measured frames, we recover multiple color frames, and these multiple frames capture variable focus depths.

\begin{figure}[t]
\begin{center}
   \includegraphics[width=1\linewidth,height = 4cm]{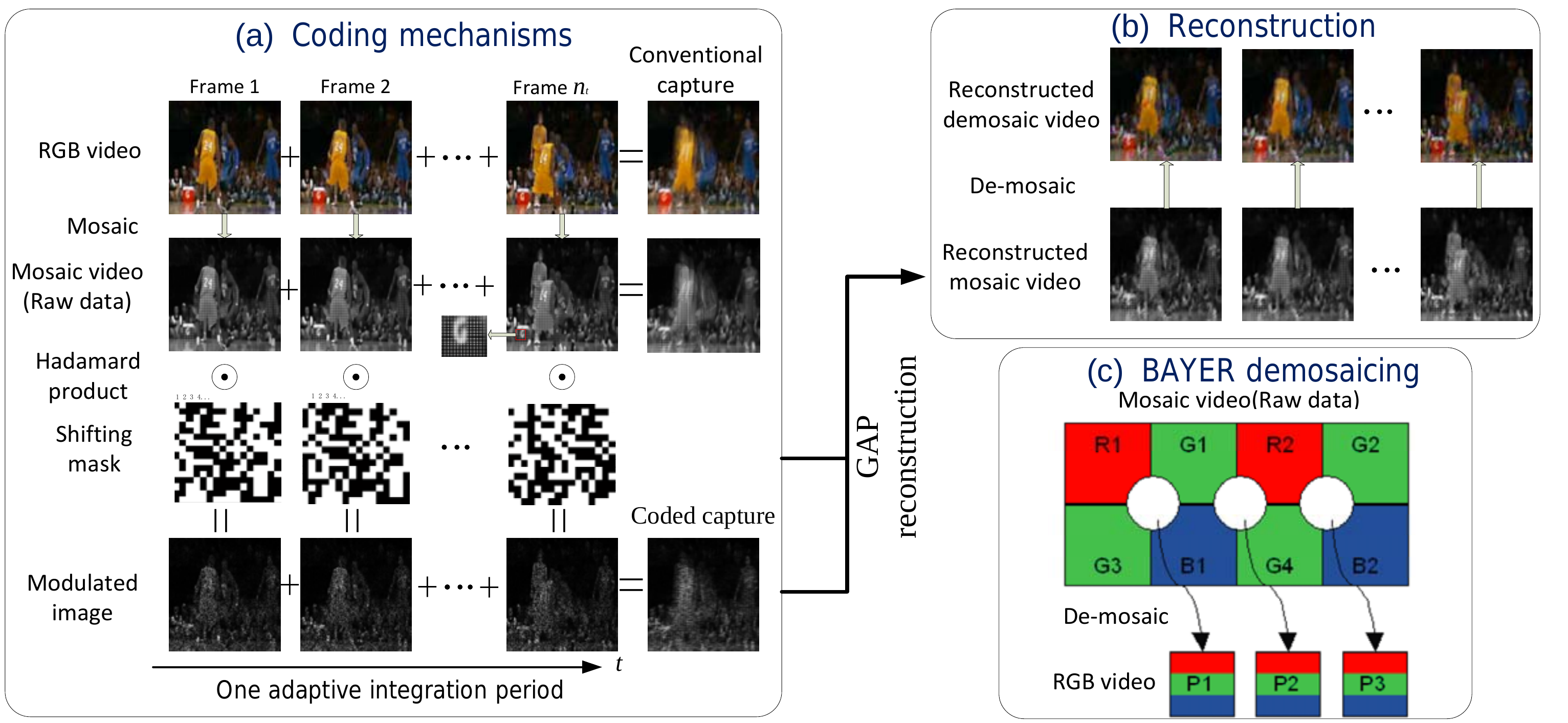}
\end{center}
\vspace{-3mm}
\caption{\small
{Illustration of the proposed method.} (a) First row shows $n_{t}$ color (RGB) frames of the original high-speed video;
second shows each color frame rearranged into a Bayer-filter mosaic;
third row depicts the (horizontally) moving mask used to modulate the high-speed frames (black is zero, white is one); fast translation manifested by a pizeoelectronic translator.
Fourth row shows the modulated frames, whose sum gives a single coded capture.
(b) Recovered RGB frames arranged into a Bayer-filter mosaic (second row), which is de-mosaicked to give the color frames (first row). (c) The demosaicing process \cite{Guppy2012}. }
\label{Fig:dec}
\vspace{-5mm}
\end{figure}

We here show example results that summarize the three key aspects of the approach: mosaicing for color, high-speed coding for video, and fast time-dependent focus for depth, with the data measured at a low frame rate. We first consider mosaicing and high-speed coding, with the focus held constant. In Figure \ref{Fig:real_ball} two compressive measurements (real data from our camera) are shown at left. These are two consecutive frames collected at frame rate 30 frames/sec, using an off-the-shelf Guppy Pro camera \cite{Guppy2012}, with a high-speed coding element, as summarized in Figure~\ref{Fig:dec}. At right in Figure \ref{Fig:real_ball}, are shown $n_t=22$ recovered frames from each of the compressive measurements: 22 color video frames recovered from a single {\em monochromatic} coded image. 
Each measurement in Figure \ref{Fig:real_ball} employs a high-speed code (here a shifting mask) to modulate the light during the integration time-period $\Delta_t$; see Figure~\ref{Fig:dec}(a).

\begin{figure}[ht!]
\begin{center}
   \includegraphics[scale=0.2]{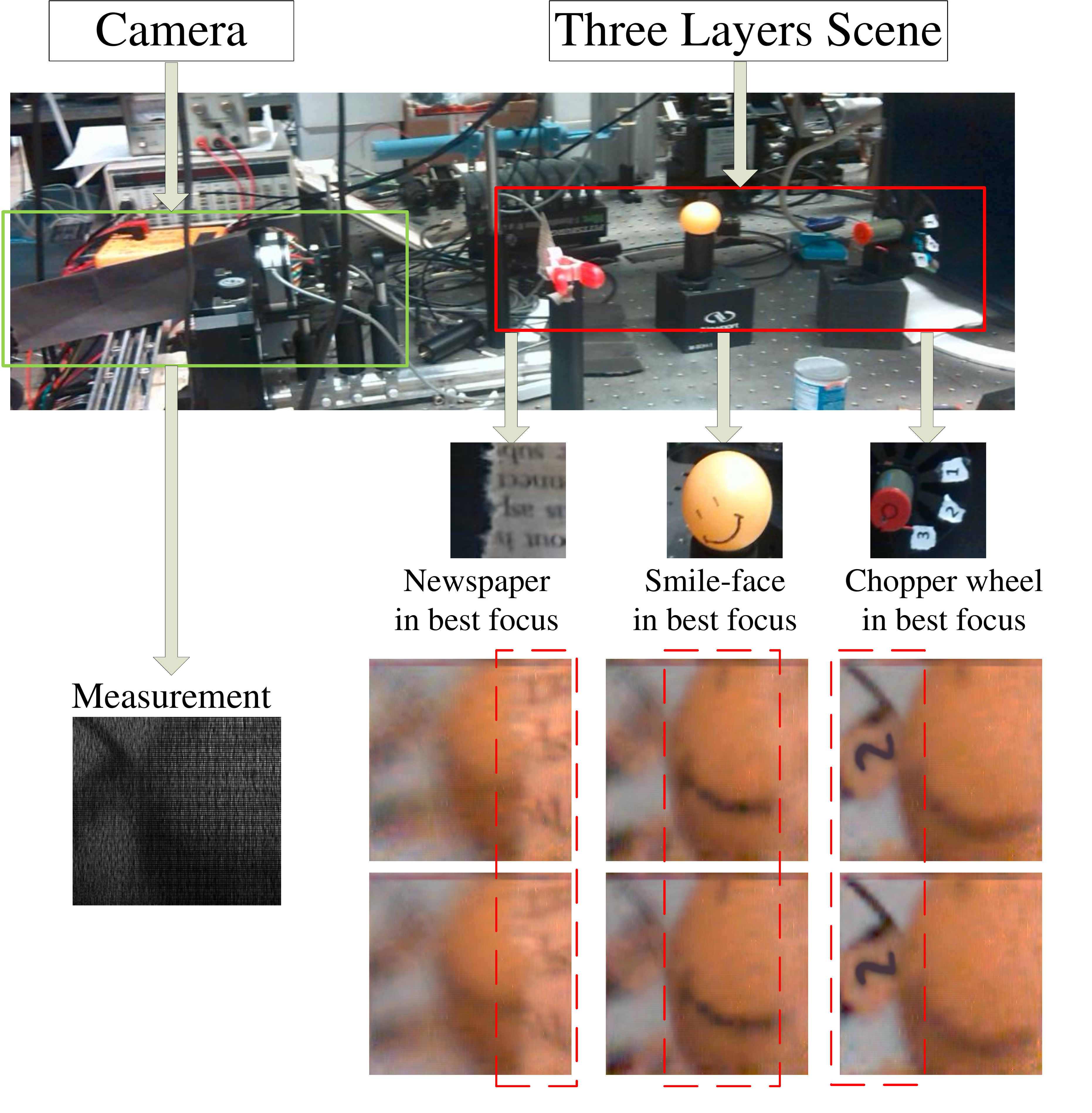}
\end{center}
\vspace{-0.5cm}
\caption{\small Experimental scene (top), three objects are placed at three depths. 6 selected recovered frames (bottom right) reconstructed all from the real single measurement (bottom left) are shown as examples.
The focal plane varies from the newspaper (near) to the chopper-wheel (far) during the integration time period. Note how chopper-wheel goes from blur (two left columns) to sharp (refer to the number ``2"). Note the motion of the \emph{moving} chopper-wheel (video online at \cite{Website}).}
\label{Fig:results_dynamic}
\vspace{-0.5cm}
\end{figure}

\begin{figure}[t]
\begin{center}
   \includegraphics[scale=0.25]{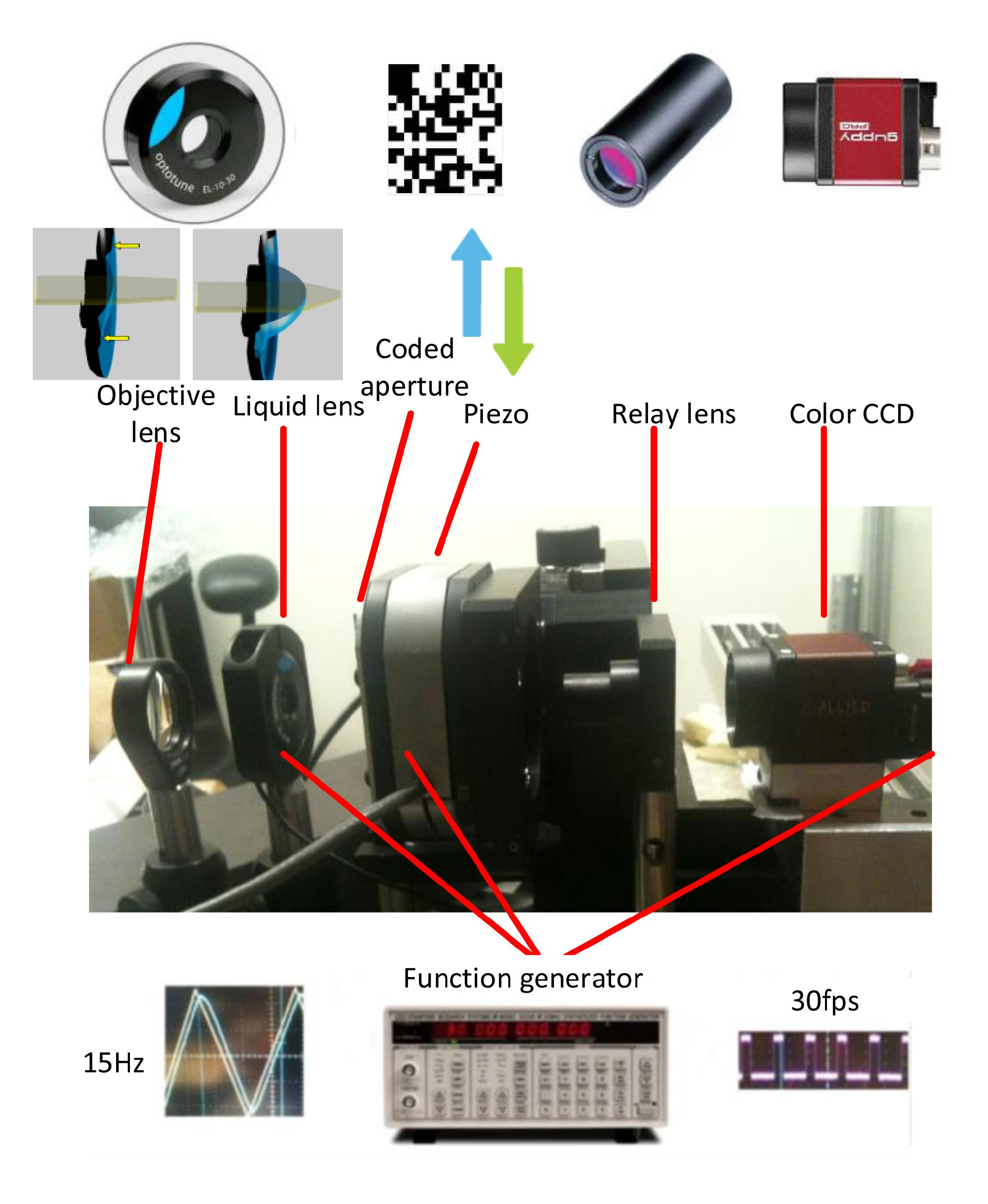}
\end{center}
\vspace{-6mm}
\caption{\small Setup of the camera.}
\label{Fig:Setup}
\vspace{-4mm}
\end{figure}

In Figure \ref{Fig:results_dynamic} we now consider results in which the focus (observed depth) is varied at a rate that is fast relative to the rate of the video camera collecting the data (now the measurements employ mosaicing, high-speed coding, and variation of the focal depth). The variable focus is manifested by varying the voltage on a liquid lens (see Figure \ref{Fig:Setup}). The coded data, with subsequent CS inversion, allows recovery of multiple frames for each measured frame. Since the focus has been adjusted at a fast rate, these high-frequency recovered frames also capture multiple depths. 

In Figure \ref{Fig:results_dynamic}, the top figure depicts the camera and scene, composed of multiple objects at varying depths. At the bottom in this figure, we depict one of the measured gray-scale frames (here measured at a frame rate of 30 frames/sec), and at bottom right is depicted the recovered color video from this single frame. Note that because of the high-speed time-varying focus, we effectively recover multiple depths, defined by the focus for which a given region of the image is sharpest.

In this paper, we describe in detail how the camera that took these measurements was implemented, with a summary provided in Figure \ref{Fig:dec} for the coding and mosaicing components, and in Figure \ref{Fig:Setup} for the camera setup. We also {discuss} a new CS inversion algorithm, that is endowed with a guarantee that upon every iteration the residual between the true video and the estimate is assured to be reduced (with technical requirements, that are discussed). This is an ``anytime'' algorithm, in that the estimated underlying video may be approximated at any time based upon computations thus far, and the quality of the inversion is guaranteed to improve with additional computations.

{The contributions of this paper are:
$i$) development of a new {\em low-cost and low-power} CS camera, that for each low-frame-rate measured image allows recovery of multiple color frames focused at different depths (at high-frame-rate); and  
$ii$) application of an anytime CS inversion algorithm to data measured by the camera, providing fast recovery of {\em high-speed motion, color and depth} information from the scene.}

\begin{figure}[t]
\begin{center}
   \includegraphics[width=1\linewidth,height = 3.0cm]{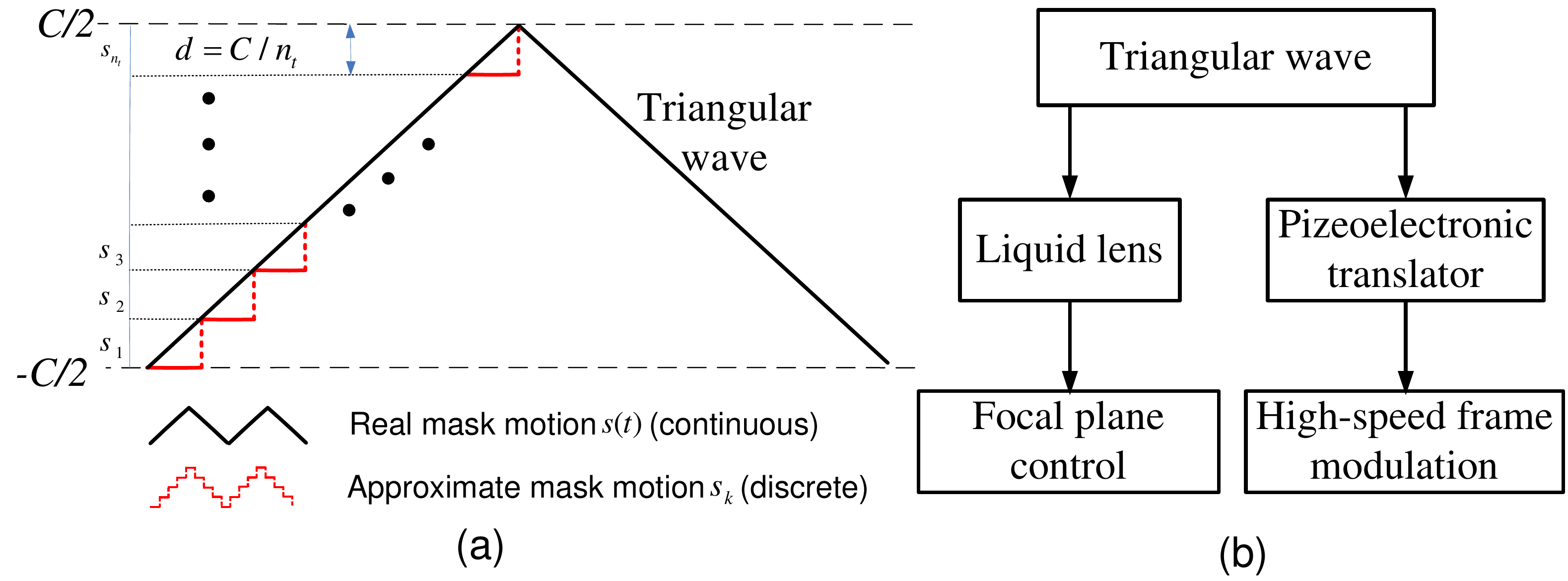}
\end{center}
\vspace{-4mm}
\caption{\small Control signal: (a) triangular wave and (b) distributed control.}
\label{Fig:mask}
\vspace{-5mm}
\end{figure}

%

\vspace{-3mm}
\section{Hardware Design}\label{Sec:Hardware}
\vspace{-2mm}
The proposed imaging system is built with an off-the-shelf camera, specifically a Guppy Pro camera \cite{Guppy2012}, by adding a {\em liquid} lens to change the focal plane, and by moving a {\em single mask} to modulate the high speed frames during one integration time-period, {both extremely low-power additions (in contrast with common alternatives in the literature, as detailed in Section \ref{sec:related})}.
The main challenge is to synchronize temporal modulation (coding) with time variation of the focus location (for capture of variable depth).
Figure \ref{Fig:Setup} depicts the setup of our camera, and Figure \ref{Fig:mask}(b) shows the synchronized control of the mask and liquid lens. 

\vspace{-2mm}
\subsection{Coding strategy}
\vspace{-2mm}
The focal plane of the liquid lens used in the camera is controlled by the voltage of the input signal, which also controls the pizeoelectronic translator to shift the mask.
The control signal (a triangular wave) is generated by a function generator and then we use power divider to distribute the signal (Figure \ref{Fig:mask}) to the liquid lens and the pizeoelectronic translator to achieve the synchronization.
The shifting (through the pizeoelectronic translator) of the same mask (Figure~\ref{Fig:dec}(a)) is utilized to modulate the high-speed frames.
This modulation enjoys advantages of low-power ($\sim 0.2W$), low-cost and low-bandwidth implementation,
compared {for example} with the modulation by liquid-crystal-on-silicon (LCoS) in \cite{Hitomi11ICCV,Reddy11CVPR} (power $>3W$, and high-bandwidth electronic switching/coding). 
In the experiments, we show that the proposed efficient coding mechanism yields similar results to that of the {relatively expensive
} LCoS-type coding.
During the calibration, we approximate the continuous moving of the mask by discrete patterns (Figure~\ref{Fig:mask}(a)).

We record temporally (and depth) compressed measurements for RGB colors on a Bayer-filter mosaic, where the three colors are arranged in the pattern shown in the right bottom of Figure \ref{Fig:dec}. The single coded image is partitioned into four components, one for R and B and two for G (each is $1/4$ the size of the original spatial image). The CS recovery (video from a single measurement) is performed separately on these four mosaiced components, and then the demosaicing process in Figure \ref{Fig:dec}(b) is employed to manifest the final color video. One may also jointly perform CS inversion on all 4 components, with the hope of sharing information on the importance of (here wavelet and DCT) components; this was also done, and the results were very similar to processing R, B, G1 and G2 separately.


\vspace{-3mm}
\subsection{Measurement model}
\vspace{-2mm}
Let $f(x,y,z,t)$ denote the continuous/analog spatio-temporal volume of the video being measured, with $(x,y,z)$ symbolizing the 3D space coordinate and $t$ denoting time.
Note the depth of the scene is defined by $z$.
Additionally, let object-space and image-space coordinates be respectively designated with unprimed and primed variables.  Given an $n_x\times n_y$-pixel sensor with pixel size $\Delta$ and integration time $\Delta_t$, space-time compressive measurements $g(x',y',t')\in \mathbb{R}^2$ are formed on the detector, with $t' < t$.  The digital data used to represent the scene consists of discrete samples of the continuous transformation
\vspace{-1mm}
\begin{eqnarray}
g(x',y',t') &\hspace{-3mm}= \int\int\int\int f(x,y,z,t)T(x\!-\!r(t),y\!-\!s(t))  \nonumber\\
&\hspace{-3mm}\times{\rm rect}\left(\frac{x-x'}{\Delta},\frac{y-y'}{\Delta}\right){\rm rect}\left(\frac{t-t'}{\Delta_t}\right)
dydxdtdz,
\label{eqn:continuous}
\vspace{-2mm}
\end{eqnarray}
\noindent where $T(x-r(t),y-s(t))$ represents a random binary transmission pattern that translates with periodic transverse $(x,y)$ motion parameterized by $(r(t),s(t))$.  The spatial and temporal pixel sampling functions, ${\rm rect}\left(\frac{x-x'}{\Delta},\frac{y-y'}{\Delta}\right)$ and ${\rm rect}\left(\frac{t-t'}{\Delta_t}\right)$, bandlimit the incident optical datastream, which is equivalently represented as $f = c\ast b$, with $\ast$ denoting the convolution operator.  Here, $c(x,y,z,t)$ denotes an instantaneous all-in-focus representation of the spatiotemporal scene and $ b(x,y,z(t))$ is a time and depth varying (blur) kernel imparted by the liquid lens on the focal volume.

The discrete formulation of the model can be simplified.
The frame at each depth ($n_t$ depths in Figure~\ref{Fig:mask}(a)) is approximated as being modulated by a single unique code (approximated by the shifting mask, Figure~\ref{Fig:mask}(a)); in reality the code/mask is always moving continuously. At each depth, the (physical/continuous) frame can be denoted by $\tilde{f}(x,y,t)$ ($z$ is now manifested by $t$), and after digitization, we represent it as ${\boldsymbol X}_k \in {\mathbb R}^{n_x\times n_y}, ~\forall k=1,\dots, n_t$.
Denoting the coding pattern of the mask by ${\boldsymbol H}_k \in {\mathbb R}^{n_x\times n_y}$, the measurement $\boldsymbol{Y} \in {\mathbb R}^{n_x\times n_y}$ is $y_{i,j} = \sum_{k=1}^{n_t} x_{i,j,k} h_{i,j,k} + e_{i,j}, \forall i=1,\dots,n_x;~ j=1,\dots,n_y;$ or
\vspace{-3mm}
\begin{equation}
\vspace{-2mm}
{\boldsymbol{Y}} =  \sum_k {\boldsymbol X}_k \odot {\boldsymbol H}_k + {\boldsymbol{E}}, 
\label{Eq:ysum}
\end{equation} 
where $\boldsymbol{E}$ denotes the noise and $\odot$ symbolizes the Hadamard (element-wise) product.
By vectorizing each frame and then concatenating them, we have
${\boldsymbol{x}} = [({\rm vec}(\boldsymbol{X}_1))^T,\dots,({\rm vec}(\boldsymbol{X}_{n_t}))^T]^T \in {\mathbb R}^{n_xn_yn_t}$, and (\ref{Eq:ysum}) can be written as:
\vspace{-4mm}
\begin{eqnarray}
\vspace{-3mm}
\boldsymbol{y} &=& [{\rm diag}({\rm vec}({\boldsymbol H}_1)),\dots, {\rm diag}({\rm vec}({\boldsymbol H}_{n_t}))] {\boldsymbol{x}} +\boldsymbol{e} \nonumber\\
&=& \boldsymbol{\Psi x} +\boldsymbol{e},\vspace{-7mm} \label{Eq:ypsix} 
\end{eqnarray}
where $\boldsymbol{\Psi} = [{\rm diag}({\rm vec}({\boldsymbol H}_1)),\dots, {\rm diag}({\rm vec}({\boldsymbol H}_{n_t}))]$.
The imaging process may therefore be modeled as in the standard CS problem.
The goal is to estimate $\boldsymbol{x}$, given ${\boldsymbol{y}}$ and $\boldsymbol{\Psi}$.
{Before presenting how we address this, we further comment on related work, now in that we have introduced the proposed hardware.}

\vspace{-2mm}
\subsection{Related work\label{sec:related}}
\vspace{-2mm}
Video compressive sensing has been investigated in
\cite{Hitomi11ICCV,HollowayICCP12,Patrick13OE,Reddy11CVPR,Veeraraghavan11TPAMI}, by capturing low frame-rate video to reconstruct high frame-rate video.
The LCoS used in \cite{Hitomi11ICCV,Reddy11CVPR} can modulate as fast as $3000$ fps by pre-storing the exposure codes, but, because the coding pattern is continuously changed at each pixel throughout the exposure, it requires considerable energy consumption ($>3W$) and bandwidth compared with the  proposed modulation, in which a single mask is translated using a pizeoelectronic translator (requiring $\sim 0.2W$). 
Similar coding was used in \cite{Patrick13OE}. However, we investigate color video here, and thus demosaicing is needed; because of the R, G and B channels, we need to properly align (in hardware of course) the mask more accurately compared with the monochromatic video in \cite{Patrick13OE}. 
Therefore, \cite{Patrick13OE} can be seen as a special case of the proposed camera.
Furthermore, we also extract the depth information from the defocus phenomenon of the reconstructed frames, which has not been considered in the above papers.

Coded apertures have been used often in computational imaging for depth estimation \cite{Levin10,Levin07,Zhou09a}. However, {these only} consider still images. From the algorithms investigated therein, one can get the depth map from a still image. 
In \cite{Kuthirummal10} an imaging system was presented that enables one to control the depth of field by varying the position and/or orientation of the image detector, during the integration time of a single photograph. However, moving the detector costs more energy than controlling the liquid lens in the proposed design (almost no power consumption), and the camera developed in \cite{Kuthirummal10} can only provide a single all-in-focus image without the depth information. 
Furthermore, no motion information is considered in the above coded-aperture cameras, while here we consider video (allowing depth estimation on moving scenes).


\vspace{-3mm}
\section{Reconstruction algorithm} \label{Sec:Algorithm}
\vspace{-3mm}
We reconstruct high-frame-rate video from low-frame-rate measurements via an {\em anytime} algorithm, the generalized alternating projection (or GAP) algorithm, first developed in \cite{GAP} {for other applications}. GAP produces a sequence of partial solutions that \emph{monotonically} converge to the \emph{true signal} (thus, anytime).
In \cite{GAP}, the authors did not mention how to select group weights and no real data or application was considered. The manner in which the GAP algorithm is employed here, as well as the application considered, is significantly different from \cite{GAP}. 
In the following, we first review the {underlying} GAP algorithm and then show how to improve it to get better results for the data considered here.

GAP is used to investigate the {\em group-sparsity} of wavelet/DCT coefficients of the video to be reconstructed. 
Let $\mathbf{T}_{x}\in\mathbb{R}^{n_{x}\times{}n_{x}}$, $\mathbf{T}_{y}\in\mathbb{R}^{n_{y}\times{}n_{y}}$,  $\mathbf{T}_{t}\in\mathbb{R}^{n_{t}\times{}n_{t}}$  be orthonormal matrices defining bases such as wavelets or DCT \cite{Mallat}. Define
$\boldsymbol{w} = \left(\mathbf{T}_{t}^{T}\otimes\mathbf{T}_{y}^{T}\otimes\mathbf{T}_{x}^{T}\right){\boldsymbol{x}}$, and  
$\bm{\Phi} =\boldsymbol{\Psi}\left(\mathbf{T}_{t}\otimes\mathbf{T}_{y}\otimes\mathbf{T}_{x}\right)$,
where $\otimes$ denotes Kronecker product \cite{Mallat}. Then we can write (\ref{Eq:ypsix}) concisely as $\boldsymbol{y}=\bm{\Phi}\boldsymbol{w}+ \boldsymbol{e}$, where $\bm\Phi\in\mathbb{R}^{n_{x}n_{y}\times n_{x}n_{y}n_{t}}$ with $\bm\Phi\bm\Phi^{T}=\mathrm{diag}\left(\mathrm{vec}\left(\sum_{k=1}^{n_{t}}{\boldsymbol{H}}_{k}\odot{\boldsymbol{H}}_{k}\right)\right)$. 
For simplification, from now we ignore possible noise $\boldsymbol{e}$.
Note that $\boldsymbol{y}$ reflects one $n_x\times n_y$ compressively measured image, as denoted at left in Figure \ref{Fig:real_ball}, and $\boldsymbol{x}=\left(\mathbf{T}_{t}\otimes\mathbf{T}_{y}\otimes\mathbf{T}_{x}\right)\boldsymbol{w}$ is the $n_x\times n_y\times n_t$ video we wish to recover (Figure \ref{Fig:real_ball} right, for $n_t=22$).

\begin{figure*}[htb]
\begin{minipage}[b]{.3\linewidth}
  \centering
  \vspace{-3mm}
  \centerline{\includegraphics[width=.85\linewidth,height = 4.0cm ]{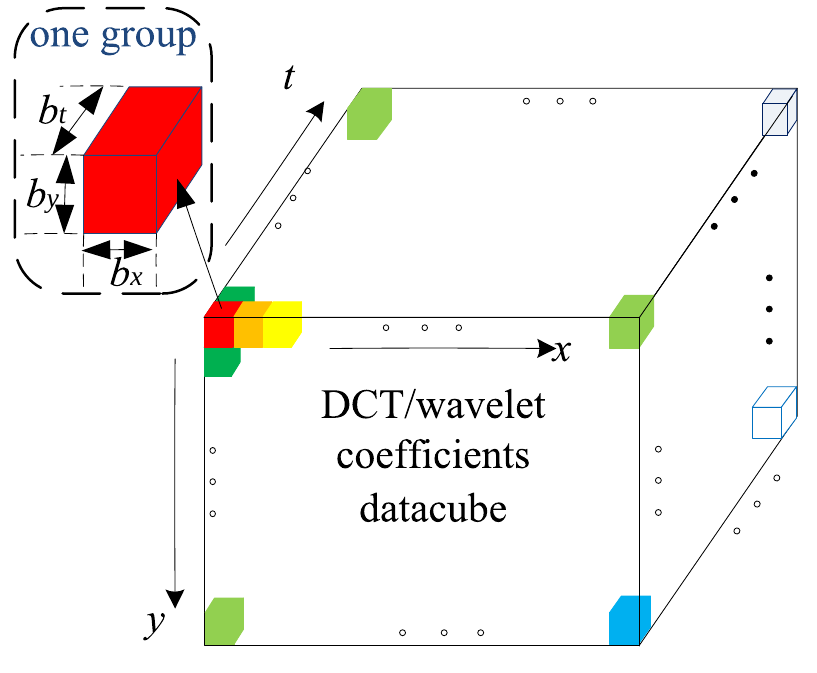}}
  \centerline{(a)}\medskip
\end{minipage}
\hfill
\begin{minipage}[b]{.3\linewidth}
  \centering
  \vspace{-3mm}
 \centerline{\includegraphics[width=0.9\linewidth,height = 3.0cm]{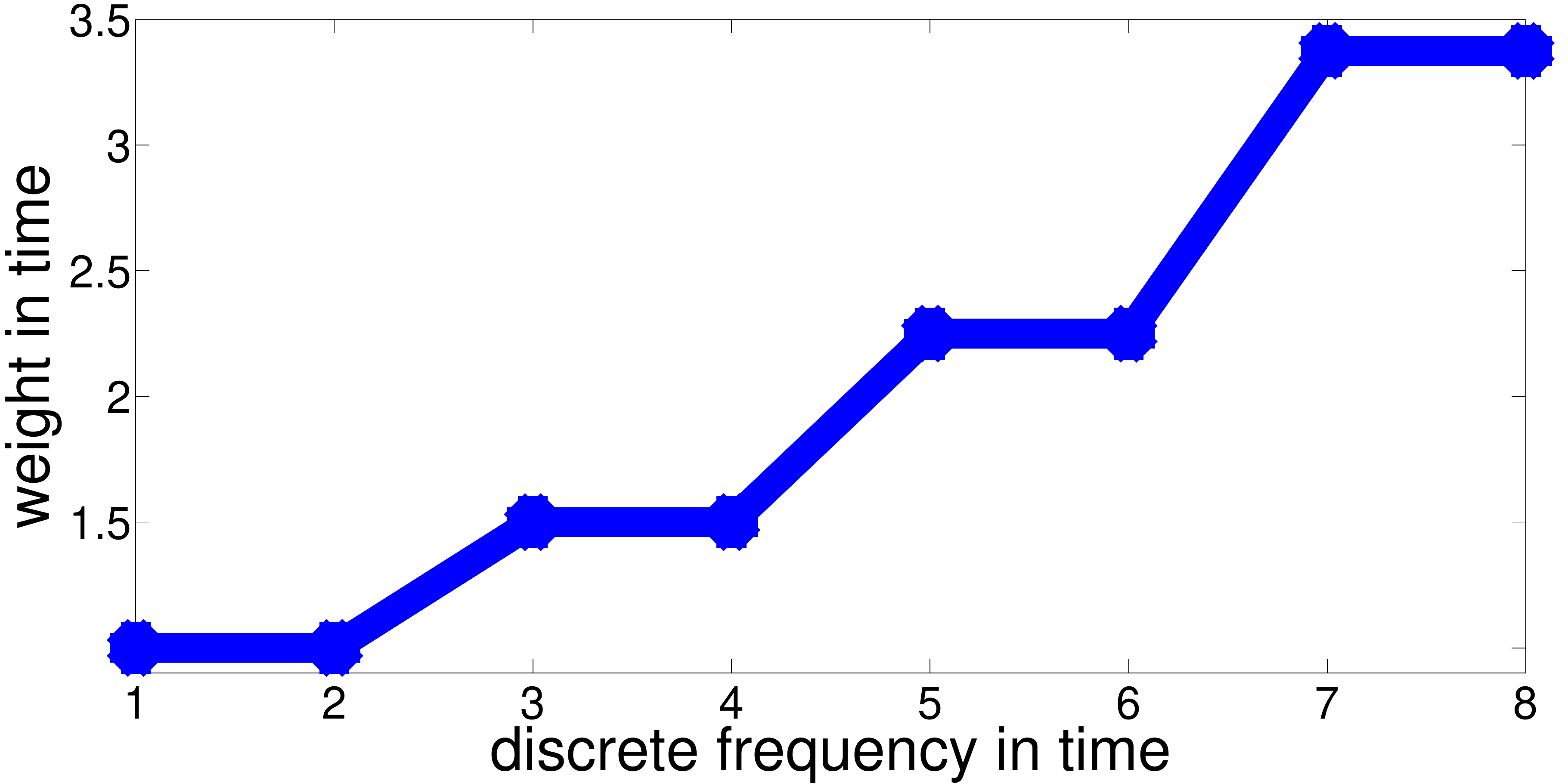}}
  \centerline{(b)}\medskip
\end{minipage}
\hfill
\begin{minipage}[b]{.3\linewidth}
  \centering
  \vspace{-3mm}
 \centerline{\includegraphics[width=1.1\linewidth,height = 4.1cm]{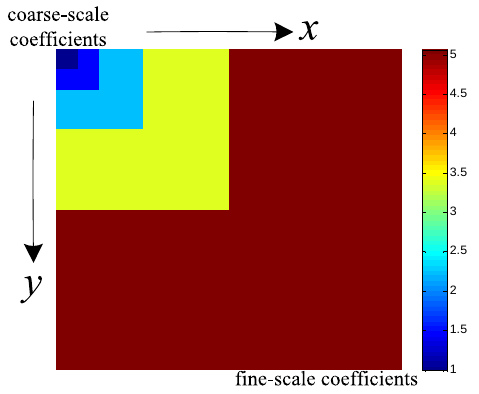}}
  \centerline{(c)}\medskip
\end{minipage}
\vspace{-3mm}
\caption{\small{(a) Demonstration of groups of the wavelet/DCT coefficients of the video. The large 3D cube represents the wavelet/DCT coefficients and the small 3D cubes in different color denote different groups, with the size $\{b_x,b_y,b_t\}$ shown by one example group on the top-left.
  (b) DCT block structure time-weight. Each $b_t$ (here $b_t=2$) frames in time share the same weight. 
  (c) Wavelet-tree structure scale-weight in space. The groups in the same wavelet level (shown in the same color) share the same weight.}\vspace{-6mm}}
\label{Fig:weight}
\end{figure*}

\vspace{-2mm}
\subsection{GAP for CS inversion}\label{sec:GAP}
\vspace{-1mm}
Let $\bm\Phi\in\mathbb{R}^{n_{1}\times{}n}$, $\boldsymbol{w}\in\mathbb{R}^{n}$, and $\boldsymbol{y}\in\mathbb{R}^{n_{1}}$, with $n_{1}<n$. Assume $\bm\Phi$ has full row rank. Let $\mathcal{G}=\{\mathcal{G}_{1}, \mathcal{G}_{2}, \cdots, \mathcal{G}_{m}\}$ be a set of nonempty mutually-disjoint and collectively exhaustive subsets of $\{1,2,\cdots,n\}$. 
Let $\bm\beta=[\beta_{1},\cdots,\beta_{m}]^{T}$ be a column of constant positive weights with $\beta_{k}$ associated with $\mathcal{G}_{k}$. We solve the weighted-$\ell_{2,1}$ minimization problem
\vspace{-3mm}
\begin{equation}\label{eq:w-L21-min-formulation}
\vspace{-1mm}
\displaystyle\min_{\,\boldsymbol{w}}\displaystyle\|\boldsymbol{w}\|_{\ell_{2,1}^{\mathcal{G}\beta}},~~\textrm{subject to}~\bm\Phi\boldsymbol{w}=\boldsymbol{y},
\vspace{-1mm}
\end{equation}
with $\|\boldsymbol{w}\|_{\ell_{2,1}^{\mathcal{G}\beta}}=\sum_{k=1}^{m}\beta_{k}\|\boldsymbol{w}_{\mathcal{G}_{k}}\|_{2}$,
where $\boldsymbol{w}_{\mathcal{G}_{k}}$ is a sub-vector of $\boldsymbol{w}$ containing components indexed by $\mathcal{G}_{k}$, and $\|\cdot\|_{2}$ denotes $\ell_{2}$ norm; $\|\cdot\|_{\ell_{2,1}^{\mathcal{G}\beta}}$ is referred as a weighted-$\ell_{2,1}$ norm or $\ell_{2,1}^{\mathcal{G}\beta}$ norm. The groups and weights are below related to the anticipated importance of wavelet/DCT coefficients; the larger $\beta_k$, the more importance is placed on the $k$th group of coefficients being sparse.

The problem in (\ref{eq:w-L21-min-formulation}) can be equivalently rewritten as
\vspace{-3mm}
\begin{equation}\label{eq:w-L21-min-formulation-2}
\vspace{-1mm}
\displaystyle\min_{\boldsymbol{w},C}C ~~\textrm{subject to}~~\|\boldsymbol{w}\|_{\ell_{2,1}^{\mathcal{G}\beta}}\leq{}C~~\mathrm{and}~~\bm\Phi\boldsymbol{w}=\boldsymbol{y}
\vspace{-3mm}
\end{equation}
Denote $B_{2,1}^{\mathcal{G}\beta}(C)=\{\boldsymbol{w}:\|\boldsymbol{w}\|_{\ell_{2,1}^{\mathcal{G}\beta}}\leq{}C\}$ and $\mathcal{S}_{\bm\Phi,\boldsymbol{y}}=\{\boldsymbol{w}:\bm\Phi\boldsymbol{w}=\boldsymbol{y}\}$, where $\mathcal{S}_{\bm\Phi,\boldsymbol{y}}$ is a given linear manifold and $B_{2,1}^{\mathcal{G}\beta}(C)$ is a weighted-$\ell_{2,1}$ ball with radius $C$. Geometrically, the problem in (\ref{eq:w-L21-min-formulation-2}) is to find the smallest weighted-$\ell_{2,1}$ ball that has a nonempty intersection with the given linear manifold; we refer to this ball as the \emph{critical ball} and denote its radius as $C^{*}$. When the smallest intersection is a singleton, the solution to (\ref{eq:w-L21-min-formulation-2}) is unique.

We solve (\ref{eq:w-L21-min-formulation-2}) as a series of alternating projection problems,
\vspace{-5mm}
\begin{eqnarray}\label{eq:w-L21-min-dist-given-C}
\vspace{-2mm}
&\displaystyle\left(\boldsymbol{w}^{(t)},\bm\theta^{(t)}\right)&=\mathrm{arg}\min_{\boldsymbol{w},\bm\theta}\displaystyle\|\boldsymbol{w}-\bm\theta\|_{2}^{2},\nonumber \\
&\textrm{subject to}& \|\bm\theta\|_{\ell_{2,1}^{\mathcal{G}\beta}}\leq{}C^{(t)}~~\textrm{and} ~~\bm\Phi\boldsymbol{w}=\boldsymbol{y},
\vspace{-2mm}
\end{eqnarray}
where a special rule is used to update $C^{(t)}$ to ensure that $\lim_{t\rightarrow\infty}C^{(t)}=C^{*}$. For each $C^{(t)}$, we solve an equivalent problem
\vspace{-2mm}
\begin{eqnarray}\label{eq:w-L21-min-formulation-ADMM}
\vspace{-2mm}
&\displaystyle\left(\boldsymbol{w}^{(t)},\bm\theta^{(t)}\right)&=\mathrm{arg}\min_{\boldsymbol{w},\bm\theta}\displaystyle\|\boldsymbol{w}-\bm\theta\|_{2}^{2}\,+\,\lambda^{(t)}\|\bm\theta\|_{\ell_{2,1}^{\mathcal{G}\beta}}\nonumber \\
&\textrm{subject to}&\bm\Phi\boldsymbol{w}=\boldsymbol{y},
\vspace{-3mm}
\end{eqnarray}
where $\lambda^{(t)}\geq0$ is the Lagrangian multiplier uniquely associated with $C^{(t)}$. Denote by $\lambda^{*}$ the multiplier associated with $C^{*}$. It suffices to find a sequence $\{\lambda^{(t)}\}_{t\geq1}$ such that  $\lim_{t\rightarrow\infty}\lambda^{(t)}=\lambda^{*}$.

We solve (\ref{eq:w-L21-min-formulation-ADMM}) by alternately projection between $\boldsymbol{w}$ and $\bm\theta$. Given one, the other is solved analytically: $\boldsymbol{w}$ is an Euclidean projection of $\bm\theta$ on the linear manifold, while $\bm\theta$ is the result of applying group-wise shrinkage to $\boldsymbol{w}$.  An attractive property of GAP is that, by using a special rule of updating $\lambda^{(t)}$, we only need to run a single iteration of (\ref{eq:w-L21-min-formulation-ADMM}) for each $\lambda^{(t)}$ to make $\{\lambda^{(t)}\}_{t\geq1}$ converge to $\lambda^{*}$. In particular, GAP starts from $\bm\theta^{(0)}=\mathbf{0}$ and computes two sequences, $\{\bm\theta^{(t)}\}_{t\geq1}$ and $\{\boldsymbol{w}^{(t)}\}_{t\geq1}$:
\begin{eqnarray}\label{eq:proj-affine}
\vspace{-2mm}
\hspace{-4mm}\boldsymbol{w}^{(t)}&\hspace{-2mm}=&\hspace{-3mm}\bm\theta^{(t-1)}+\bm\Phi^T(\bm\Phi\bm\Phi^T)^{-1}(\boldsymbol{y}-\bm\Phi\bm\theta^{(t-1)}),
\\\label{eq:weighted-L21-proj-final-t}
\hspace{-4cm}\bm\theta^{(t)}_{\mathcal{G}_{k}}&\hspace{-2mm}=&\hspace{-3mm}\boldsymbol{w}^{(t)}_{\mathcal{G}_{k}}\,\max\left\{1-\frac{\lambda^{(t)}\beta_{k}}{\left\|\boldsymbol{w}^{(t)}_{\mathcal{G}_{k}}\right\|_{2}},\,0\right\},\,\forall k=1,\dots,m
\vspace{-2mm}
\end{eqnarray}
\vspace{-2mm}
\begin{equation}\label{eq:lambdat}
\hspace{-15mm}\text{where} ~~~\lambda^{(t)}=\left\|\boldsymbol{w}^{(t)}_{\mathcal{G}_{j^{(t)}_{m^{\star}+1}}}\right\|_{2}\beta_{j^{(t)}_{m^{\star}+1}}^{-1},~~ m^{\star}<m
\end{equation}
with $\left(j^{(t)}_{1},j^{(t)}_{2},\cdots,j^{(t)}_{m}\right)$ a permutation of $\left(1,2,\cdots,m\right)$ such that
$
\left\|\boldsymbol{w}^{(t)}_{\mathcal{G}_{j^{(t)}_{1}}}\right\|_{2}\beta_{j^{(t)}_{1}}^{-1}\,\,\geq\,\,\left\|\boldsymbol{w}^{(t)}_{\mathcal{G}_{j^{(t)}_{2}}}\right\|_{2}\beta_{j^{(t)}_{2}}^{-1}\,\,\geq\cdots\,\,\geq\,\,\left\|\boldsymbol{w}^{(t)}_{\mathcal{G}_{j^{(t)}_{m}}}\right\|_{2}\beta_{j^{(t)}_{m}}^{-1}. 
$

The algorithm (\ref{eq:proj-affine})-(\ref{eq:weighted-L21-proj-final-t}) is referred as \emph{generalized alternating projection} (or GAP) \cite{GAP} to emphasize its difference from alternating projection (AP) in the conventional sense: conventional AP produces a sequence of projections between two \emph{fixed} convex sets, while GAP produces a sequence of projections between two convex sets that undergo systematic changes over the iterations. In the GAP algorithm as shown in (\ref{eq:proj-affine})-(\ref{eq:weighted-L21-proj-final-t}), the alternating projection is performed between a fixed linear manifold $\mathcal{S}_{\bm\Phi,\boldsymbol{y}}$ and a changing weighted-$\ell_{2,1}$ ball, i.e., $B_{2,1}^{\mathcal{G}\beta}(C^{(t)})$ whose radius $C^{(t)}$ is a function of the iteration number $t$.

By interpreting $\frac{1}{\lambda^{(t)}}\|\bm\theta-\boldsymbol{w}\|_{2}^{2}$ as a penalty to enforce $\bm\theta=\boldsymbol{w}$, one may view that iteration of (\ref{eq:w-L21-min-formulation-ADMM}) with $t$ constituting a penalty method for solving the following constrained problem, 
\vspace{-3mm}
\begin{equation}\label{eq:w-L21-min-formulation-ADMM-overlap}
\vspace{-2mm}
\displaystyle\min_{\bm\theta}\|\bm\theta\|_{\ell_{2,1}^{\mathcal{B}\beta}}~~\textrm{subject to}~~
\bm\theta=\boldsymbol{w}~~\mathrm{and}~~\bm\Phi\boldsymbol{w}=\mathbf{y},
\vspace{-1mm}
\end{equation}
which is an equivalent formulation of (\ref{eq:w-L21-min-formulation}). 
The GAP algorithm in (\ref{eq:proj-affine})-(\ref{eq:lambdat}) is a special penalty method for solving (\ref{eq:w-L21-min-formulation-ADMM-overlap}), using (\ref{eq:lambdat}) to adjust the penalty strength $\left\{\lambda^{(t)}\right\}$.  
Bregman iteration~\cite{Yin08bregman} can also solve (\ref{eq:w-L21-min-formulation}) or (\ref{eq:w-L21-min-formulation-ADMM-overlap}). 
However, Bregman penalizes $\|\boldsymbol{y}- \boldsymbol{\Phi w}\|_2^2$, while GAP keeps $\boldsymbol{y}=\boldsymbol{\Phi w}$ as a constraint and fulfills it via the orthogonal projection in (\ref{eq:proj-affine}).  Under a set of group-restricted isometry property (group-RIP) conditions, the use of (\ref{eq:proj-affine})-(\ref{eq:lambdat}) ensures monotonic decrease of the reconstruction error to zero and makes GAP an anytime algorithm~\cite{GAP}. By contrast, Bregman iteration and classic penalty method (which adjusts $\lambda^{(t)}$ differently) do not have the anytime property, nor do other popular algorithms such as TwIST and ADMM~\cite{ADMM2011Boyd}. 
 

\vspace{-3mm}
\subsection{{Extension} of GAP for the proposed camera}
\vspace{-2mm}
The diagonalization of $\bm\Phi\bm\Phi^{T}$ is the key to fast GAP recovery of video.
The inversion of $\bm\Phi\bm\Phi^{T}$ in (\ref{eq:proj-affine}) now just requires computation of the reciprocals of the diagonal elements, as a result of the {\em hardware implementation} of the proposed camera.
Best results were found when $\mathbf{T}_x$ and $\mathbf{T}_y$ correspond to a wavelet basis (here the Daubechies-8 \cite{Mallat}), and $\mathbf{T}_t$ corresponds to a DCT. The basis-function weights ($\beta$) are defined with respect to these bases, and the groups are manifested in the domain of these wavelet-DCT coefficients (see Figure \ref{Fig:weight}(a) for a depiction of the groups). For $\mathbf{T}_t$ the coefficients are arranged from low frequencies to high frequencies in Figure \ref{Fig:weight}(b), and for $\mathbf{T}_x$ and $\mathbf{T}_y$ the 2D arrangement of coefficients is as is done typically with wavelets \cite{Mallat}, and illustrated in Figure \ref{Fig:weight}(c). Let $\{b_x,b_y,b_t\}$ represent the edge lengths of each group of coefficients (Figure \ref{Fig:weight}(a)). In all experiments, $b_x=b_y=2$ and $b_t = \left[\frac{n_t}{4}\right]$, where $[\hspace{0.03in}]$ denotes the closest integer of the number inside $[\hspace{0.03in}]$.
The weight is constituted by 
the product of a time-weight associated with the DCT (Figure \ref{Fig:weight}(b)) and a scale-weight associated with the wavelets (Figure \ref{Fig:weight}(c)). We show the details of the weight design in the following.
 
Let $\{g_x,g_y,g_t\}$ index each group with $g_x =1,\dots, \frac{n_x}{b_x}, ~g_y = 1,\dots,\frac{n_y}{b_y},~g_t = 1,\dots,\frac{n_t}{b_t}$.
The time weights are defined by 
${\beta}_t(g_t)= a^{(g_t-1)}$.
For the scale weights,
we exploit the wavelet-tree structure (Figure \ref{Fig:weight}(c)), and enforce the groups in the $\ell$th level (assuming $L$ levels in total, and $\ell = 1,\dots,L$) of the wavelet coefficients sharing the same weight, determined by
 $\mathcal{\beta}_{x,y}(g_x^{\ell},g_y^{\ell}) = a^{(\ell-1)}, \forall~
g_x^{\ell} = \frac{n_x^{(\ell-1)}}{b_x}+1,\dots, \frac{n_x^{\ell}}{b_x},
~g_y^{\ell} = \frac{n_y^{(\ell-1)}}{b_y}+1,\dots, \frac{n_y^{\ell}}{b_y}$,
with $\left\{n_x^{\ell}, n_y^{\ell},n_t^{\ell}\right\}$ denoting the end points of wavelet coefficients at the $\ell$th level, and $n_x^0 =n_y^0 =0$.
The weight for each 3D group is ${\beta}(g^{\ell}_x,g^{\ell}_y,g_t) = \mathcal{\beta}_{x,y}(g_x^{\ell},g_y^{\ell}){\beta}_t(g_t)$.
Setting $a=1.5$ was found to yield good results. 
After constructing the groups and weights as above, the GAP performance is improved significantly in the application here. 

\vspace{-1mm}
\subsection{Temporal overlap in inversion}
\vspace{-1mm}
In Section \ref{sec:GAP}, each coded CS measurement $\boldsymbol{Y}_l$ is employed to recover $n_t$ frames of video. This may lead to discontinuities in the video recovered from consecutive CS measurements. To mitigate this, we also consider the \emph{joint} inversion of two consecutive CS measurements, $\boldsymbol{Y}_l$ and $\boldsymbol{Y}_{l+1}$, from which $2n_t$ consecutive frames are recovered at once. Therefore the $n_t$ frames associated with $\boldsymbol{Y}_l$ are estimated jointly from $\boldsymbol{Y}_{l-1} \cup \boldsymbol{Y}_{l}$ and (separately) from $\boldsymbol{Y}_{l} \cup \boldsymbol{Y}_{l+1}$. The final recovered video within a particular contiguous set of $n_t$ frames is taken as the average of the results inferred from $\boldsymbol{Y}_{l-1} \cup \boldsymbol{Y}_{l}$ and $\boldsymbol{Y}_{l} \cup \boldsymbol{Y}_{l+1}$. As demonstrated below, this tends to improve the quality of the recovered video (yields smoother results).

\begin{figure}[htb]
\vspace{-4mm}
  \centering
  \centerline{\includegraphics[scale=0.13]{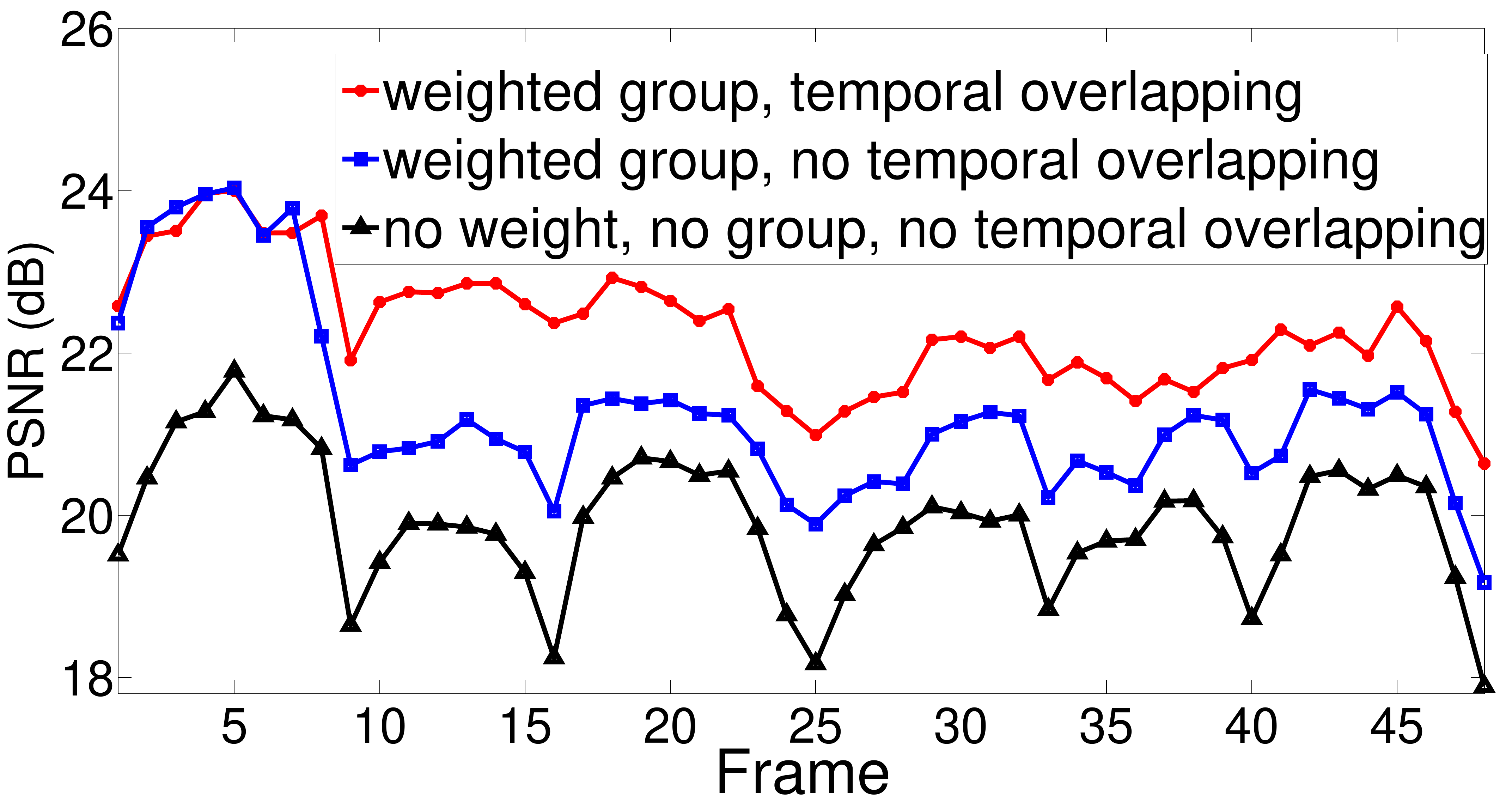}}
  \vspace{-2mm}
  \caption{\small PSNR comparison of weighted group (average PSNR: 22.13dB), no weight, no group (average PSNR: 19.92dB), and temporal overlapping weighted group (average PNSR: 22.80dB).}
  \label{Fig:Weight}
  \vspace{-4mm}
\end{figure}
  
\begin{figure}[htb]
\centering
\vspace{-3mm}
  \centerline{\includegraphics[scale=0.13]{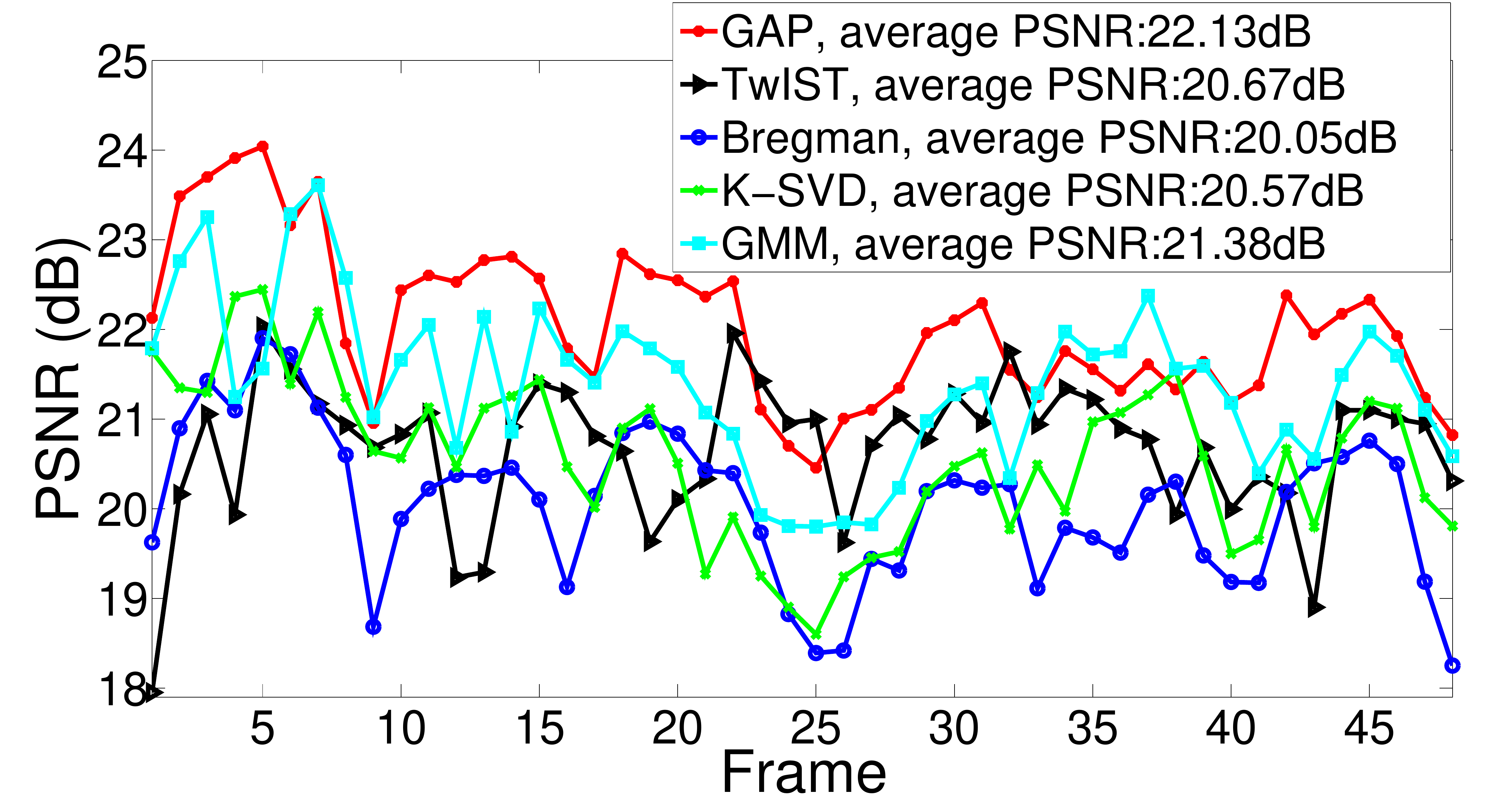}}
  \vspace{-1mm}
\caption{\small PSNR comparison of GAP, TwIST, linearized Bregman, K-SVD, and GMM algorithms with the simulation dataset.\vspace{-4mm}}
\label{Fig:Weight_Algorithm}
\vspace{-3mm}
\end{figure}

\vspace{-3mm}
\section{Simulation results} \label{Sec:Simulation}
\vspace{-2mm}
To demonstrate the performance of the reconstruction algorithm, we {start with simulated data (see Section~\ref{Sec:Real} for real data from the proposed camera), and} employ a color video sequence in which a basketball player performs a dunk \cite{Website}; this video is challenging due to the complicated motion of the basketball players and the varying lighting conditions and multiple depths scene; see the example video frames in Figure \ref{Fig:dec}(a).
We consider a binary mask, with 1/0 coding drawn at random Bernoulli(0.5); the code is shifted spatially via the coding mechanism in Figure \ref{Fig:dec}(a)), as in our physical camera. 
The video frames are $256\times 256$ spatially, and we choose $n_t=8$.
 
We first investigate the efficacy of weighted groups in the proposed GAP algorithm.
Figure \ref{Fig:Weight} demonstrates the 
improvement by the weighting and grouping of the wavelet/DCT coefficients; these parameters (weights and groups) were not optimized, and many related settings yield similar results -- there is a future opportunity for optimization. In Figure \ref{Fig:Weight} we also demonstrate the performance improvement manifested by temporal overlapping and averaging results from two consecutive measurements. 
Note that \emph{without} temporal overlapping, the PSNR degrades for frames at the beginning ($e.g.$, 1, 9, 17, ...) and end (8, 16, 24, ...) of a given measurement, while the PSNR curve with temporal overlapping (red curve) is much smoother. The experiments with (real) measured data consider temporal overlapping when performing inversion.

\begin{figure}[htb]
\begin{minipage}[b]{.48\linewidth}
  \centering
  \vspace{-3mm}
  \centerline{\includegraphics[width=1\linewidth,height = 2.5cm ]{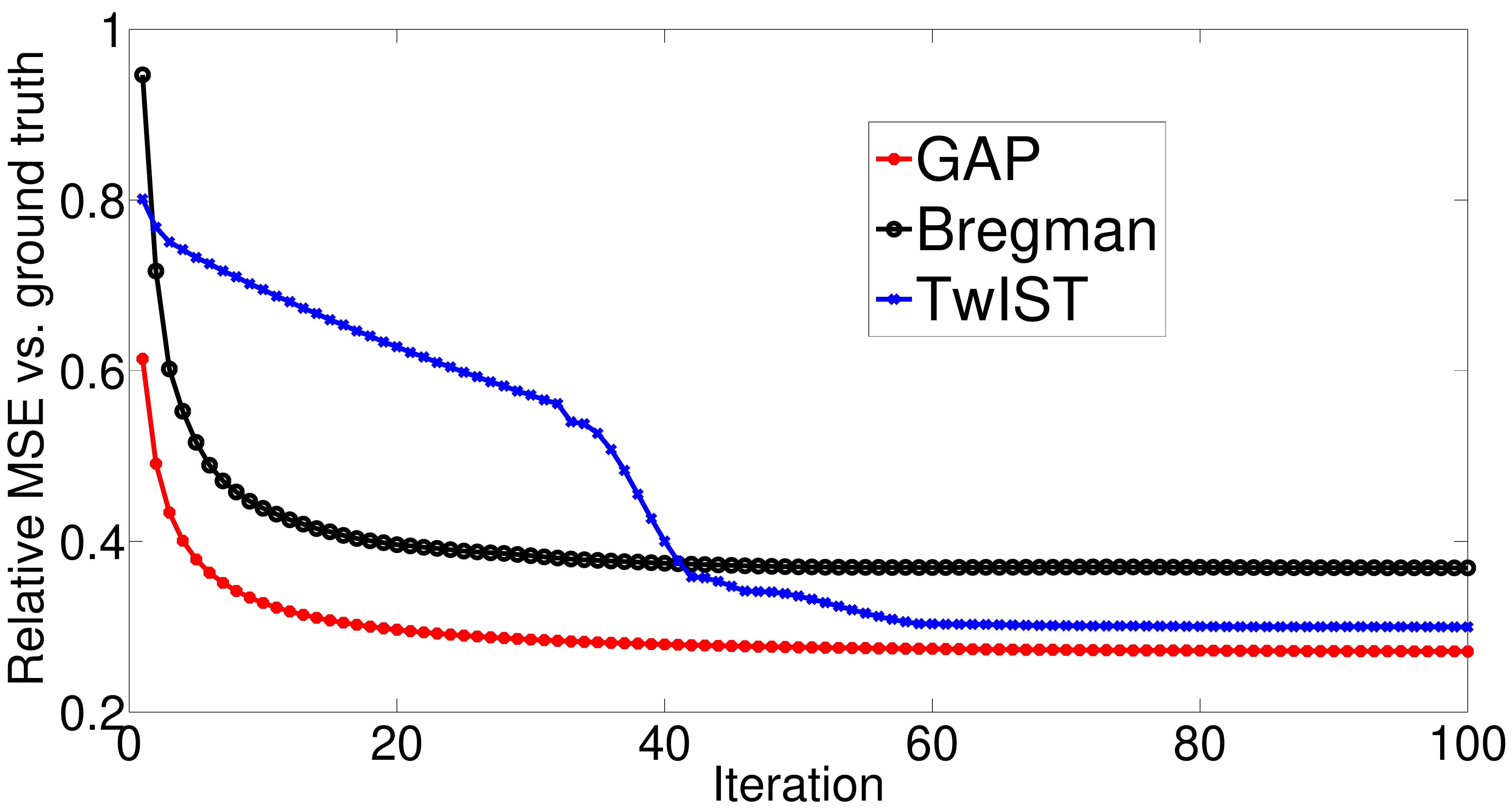}}
  \centerline{(a)}\medskip
\end{minipage}
\hfill
\begin{minipage}[b]{.50\linewidth}
  \centering
  \vspace{-3mm}
 \centerline{\includegraphics[width=1.0\linewidth,height = 2.55cm]{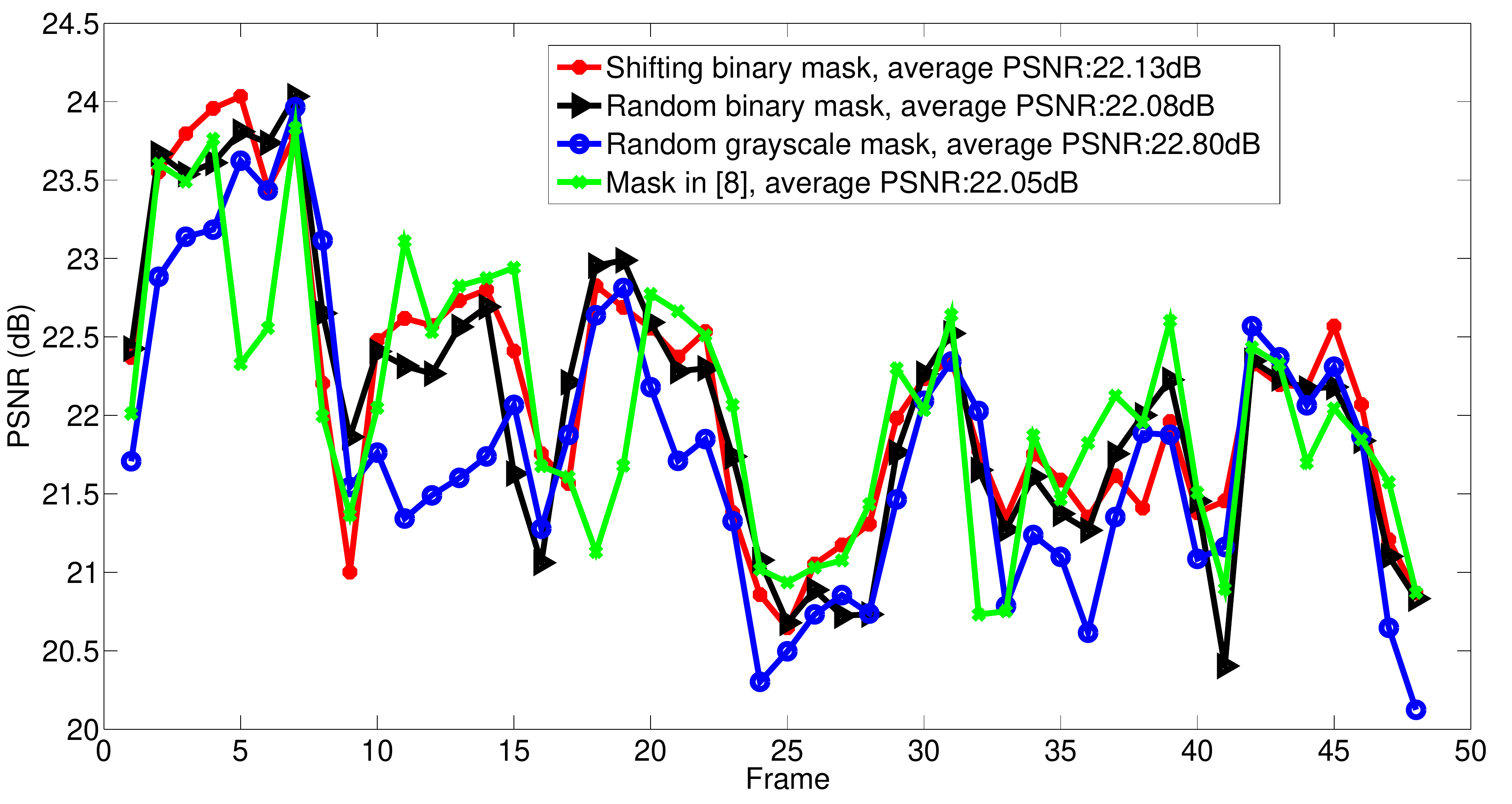}}
  \centerline{(b)}\medskip
\end{minipage}
\vspace{-3mm}
\caption{\small{(a) Convergence comparison of GAP, TwIST, and linearized Bregman.
  (b) PSNR of reconstructed video frames with three different coding scenarios. ``Shifting binary mask" means the coding mechanism proposed in our camera.
    ``Random binary mask" denotes that for every frame, each element of the mask is randomly drawn from Bernoulli$(0.5)$.
    ``Random grayscale mask" symbolizes each element of the mask is randomly drawn from $[0,1]$ for every frame. }}
\vspace{-5mm}
\label{Fig:Convergence}
\end{figure}


\vspace{-2mm}
\subsection{Reconstruction performance and convergence}
\vspace{-2mm}
The developed GAP algorithm is compared with the following: 
$i$) two-step iterative shrinkage/thresholding (TwIST) \cite{Bioucas-Dias2007TwIST} (with total variation norm), 
$ii$) K-SVD \cite{Aharon06TSP} with orthogonal matching pursuit (OMP) \cite{Tropp07ITT} used for inversion, 
$iii$) a Gaussian mixture model (GMM) based inversion algorithm \cite{Chen10SPT,Yang13GMM,Yu12IPT}, and 
$iv$) the linearized Bregman algorithm \cite{Yin08bregman}.
The $\ell_1$-norm of DCT or wavelet coefficients is adopted in linearized Bregman algorithm with the same transformation as GAP.
GMM and K-SVD are patch-based algorithms and {we used a separate dataset for training}.
A batch of training videos were used (shown in the website \cite{Website}) to pre-train K-SVD and GMM, and we selected the best reconstruction results for presentation here.
The PSNR curves of videos reconstructed by GAP and the four alternative algorithms are shown in Figure \ref{Fig:Weight_Algorithm}, demonstrating the GAP algorithm outperforms the other algorithms by 0.7-2dB higher in PSNR.
In this simulation, the temporal overlap was not used.
The readers can refer to the reconstructed videos on the website \cite{Website}.

We further investigate the convergence properties of GAP, TwIST and linearized Bregman, as they are each solved iteratively.
We run each algorithm a total of $100$ iterations
and compute the relative mean square error (MSE) of the estimate compared with the ground truth for each iteration. Relative MSE versus iteration number are plotted in Figure \ref{Fig:Convergence}(a).
It can be seen that GAP and linearized Bregman converge much faster than TwIST, while GAP provides the smallest relative MSE in every iteration.
We also verify the {\em anytime} property of GAP by computing the first-order difference of the MSE for each iteration (compared with TwIST and Bregman); see Supplementary Material \cite{Website}.

\vspace{-2mm}
\subsection{Comparison with other coding mechanisms}
\vspace{-2mm}
One advantage of the proposed coded aperture compressive camera (manifested by spatially shifting a \emph{single} binary mask), compared with others \cite{Hitomi11ICCV,Reddy11CVPR}, is its low-power, low-bandwidth characteristics. However, the use of a single shifted binary mask to yield temporal coding may appear limiting, and therefore it is of interest to compare to other strategies. 
In Figure \ref{Fig:Convergence}(b) we compare the proposed translated coding mechanism in our hardware system with even more general coding strategies than in  \cite{Hitomi11ICCV,Reddy11CVPR}. Specifically, we compare to a unique random binary mask at each time point, for which \emph{each} of the $n_t$ codes is a \emph{distinct} i.i.d. draw Bernoulli(0.5). We also consider the case for which each code/mask element, for each of the $n_t$ codes, is drawn uniformly at random from [0,1], reflecting the degree of code transmission (each of the $n_t$ codes is distinct, and each is not restricted to be binary). 
Summarizing Figure \ref{Fig:Convergence}(b), the simple shifted binary code in the proposed system yields similar results (even a little higher PSNR for some datasets) compared to these alternative coding strategies, {at an order of magnitude less power}. 
We found similar results when comparing to the coding strategies in \cite{Hitomi11ICCV,Reddy11CVPR}.

\vspace{-3mm}
\section{Experimental results: real data} \label{Sec:Real}
\vspace{-2mm}
The physical camera we have developed captures the measurements (coded capture) at $1/\Delta_t=30$ fps, and the reconstructed video has 660 fps ($n_t=22$, although different $n_t$ may be considered in the inversion). One result from this camera was shown in Figure \ref{Fig:real_ball} (recovery of high-speed motion). From the reconstructed frames in Figure \ref{Fig:real_ball}, one can clearly identify the spin of the red apple and the rebound of the yellow orange; the full video is at \cite{Website}, along with many addition examples. 
At \cite{Website}, we show comparisons to a diverse set of alternative algorithms, for example via \cite{Yin08bregman}.

All algorithms considered here were implemented in Matlab 2012(a), and the experiments are conducted on a PC with a CPU@3.30 GHz and 16GB RAM.
For the real data ($512\times 512\times 22$),
GAP, TwIST, and linearized Bregman use 50, 100, and 500 iterations, respectively (required to yield good results). Each iteration in these three algorithms are similar (around 0.8 seconds).
Hence, TwIST and linearized Bregman cost much longer ($>2\times$) time than GAP, but typically provide worse results, and do not have an anytime property.
K-SVD and GMM may be made fast if parallel computing is used (processing the multiple patches in parallel with GPU or networks), but for serial computing on a PC like that considered here
these methods are slower than GAP.

\vspace{-3mm}
\subsection{Recovery of depth and motion simultaneously}
\vspace{-2mm}
Figure \ref{Fig:results_dynamic} shows the reconstruction frames (6 out of 14 frames are shown for demonstration) recovered from one measurement.
The three objects, ``newspaper," ``smile-face" and ``chopper-wheel" are in three different depths.
The motion of the \emph{rotating} chopper-wheel is also recovered from the reconstructed frames.
The first column of the recovered frames (bottom right of Figure \ref{Fig:results_dynamic}) shows the newspaper is in focus; the second column shows the smile-face is in focus, and finally, the chopper-wheel is in focus at column three.
We have built a depth-frames relationship table with calibration. 
The newspaper is best in focus in frame 3, which corresponds to 14cm away from the objective lens (truth 15cm).
The smile-face is best in focus in frame 8, corresponding to 40cm (truth 38cm).
The chopper-wheel is best in focus in frame 12, which corresponds to 64cm (truth 65cm).
It can be seen the depths of these objects are estimated correctly.

\vspace{-3mm}
\subsection{Recovery of high-speed motion}
\begin{figure}[htbp]
\vspace{-0.7cm}
\begin{center}
   \includegraphics[scale=0.37]{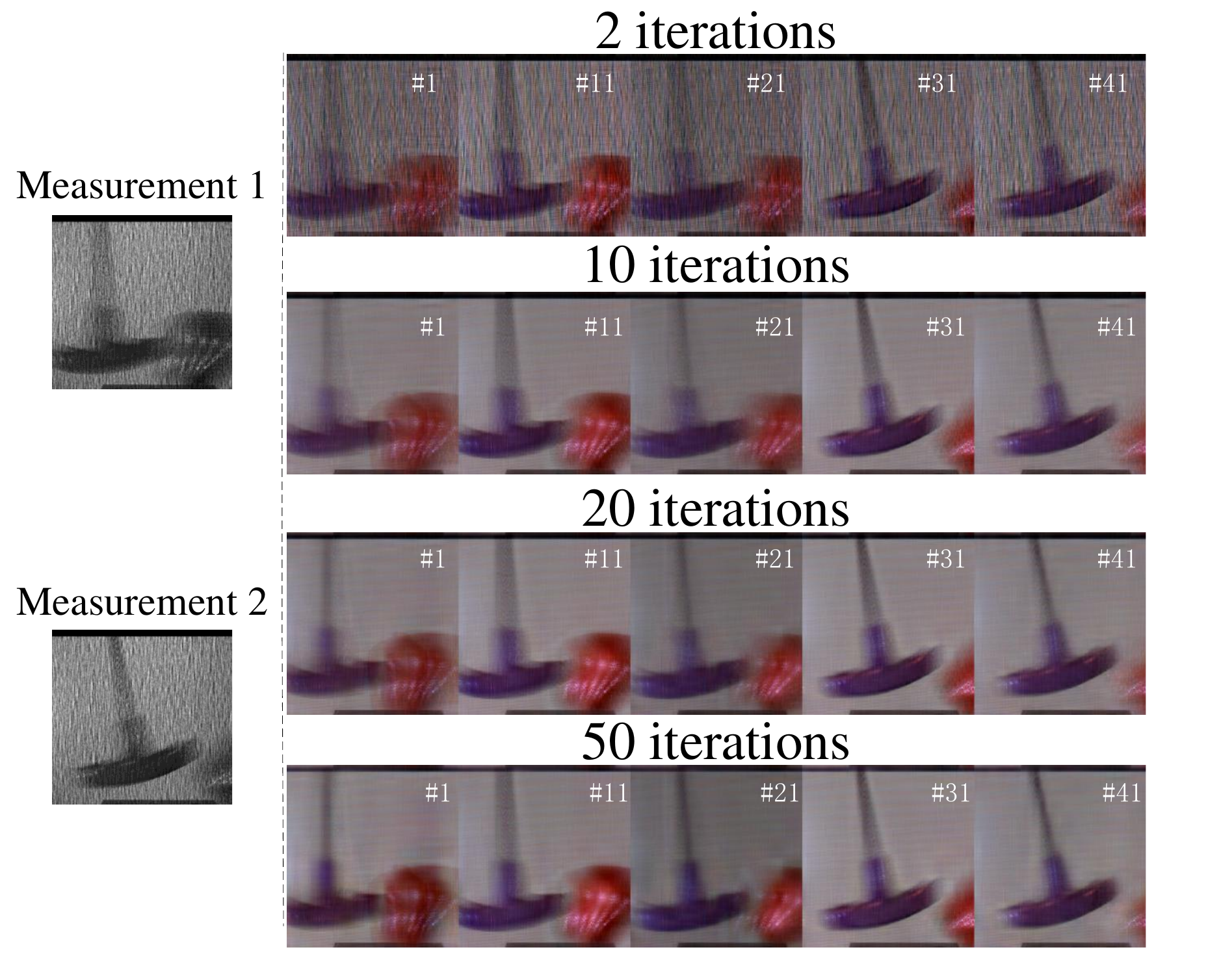}
\end{center}
\vspace{-0.5cm}
\caption{\small{Reconstruction from two real measurements (left part). 
Five selected frames out of the 44 reconstructed video frames ($n_t=22$ recovered video frames from each compressive measurement) after 2, 10, 20, and 50 iterations of GAP are shown in the right part. 
}}
\label{Fig:real_hammer}
\vspace{-0.3cm}
\end{figure}

\vspace{-3mm}
As one additional example of high-speed motion recovery (assuming all the objects are in focus, $i.e.$, here without the liquid lens), Figure \ref{Fig:real_hammer} shows the results of a purple hammer quickly hitting a red apple. In this dataset, 44 frames are reconstructed from 2 measurements (left part) showing the entire process of hitting (full video is at \cite{Website}), and
5 example frames out of 44 are plotted in the right part of Figure \ref{Fig:real_hammer}. 
To demonstrate the {\em anytime} property of GAP, we show the results after 2, 10, 20 and 50 iterations.
Note that good results are manifested with as few as 10 iterations, with convergence after about 20.

\begin{figure}[htbp]
\vspace{-0.4cm}
\begin{center}
   \includegraphics[scale=0.4]{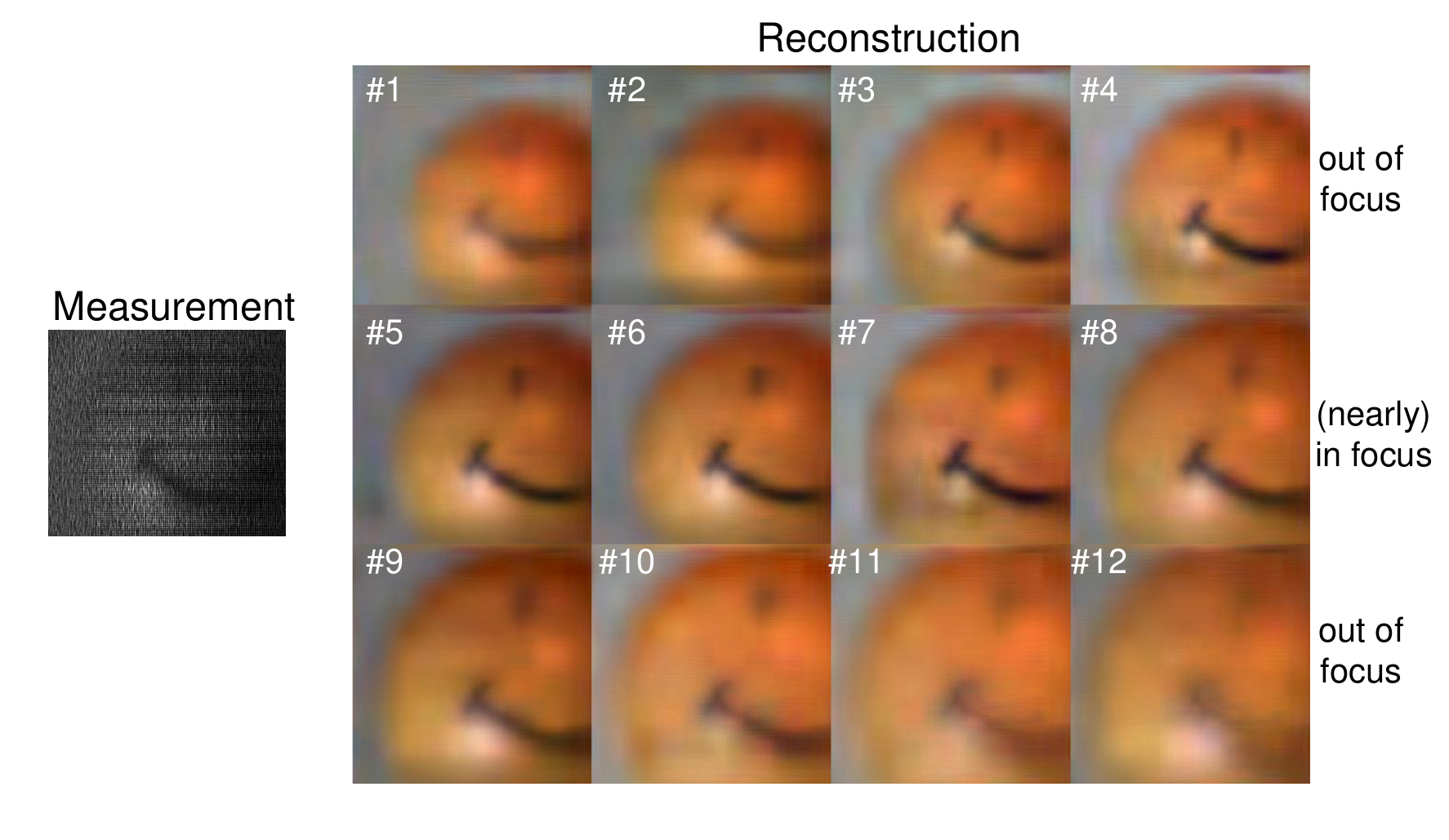}
\end{center}
\vspace{-0.5cm}
\caption{\small{Reconstruction from a real single measurement (left part). 
12 selected frames are shown in the right part to demonstrate the process of the smile-face from out-of-focus to in focus and then out-of-focus again. }}
\label{Fig:real_smile}
\vspace{-0.7cm}
\end{figure}


\vspace{-0.2cm}
\subsection{Recovery of depth}
\vspace{-0.2cm}
When the scene is not moving, we can get space-depth-color information from the reconstructed data.
An example is shown in Figure~\ref{Fig:real_smile}. We can see that the smile-face is first out-of-focus, then in-focus and finally out-of-focus again.


\vspace{-0.4cm}
\section{Conclusions} \label{Sec:Con}
\vspace{-0.3cm}
This paper proposes a means of recovering depth, time, and color information from a single coded image, via development of a new color CS camera for high-speed depth-video reconstruction.
In the presented computational time comparisons, GAP was run in MATLAB, since that was the language in which all of the comparison algorithms had available code, and therefore provided a good comparison point for \emph{relative} speed. In the context of \emph{absolute} speed, we have implemented GAP in C++ on a GPU, and the \emph{total} time for reconstructing a $512\times 512\times 22$ video from a single CS measurement is less than 0.5 seconds, {opening the door for real-time fast (3D) video capturing and reconstruction.} 
\vspace{-0.4cm}
\section*{Acknowledgements}
\vspace{-0.3cm}
This work was supported by DARPA under the KECoM program, and by the DOE, NGA, ONR, NSF and ARO. 

\vspace{-0.3cm}
{\small
\bibliographystyle{ieee}

}

\end{document}